\documentclass{article}
\usepackage[final]{nips_2016}

\usepackage[utf8]{inputenc} 
\usepackage[T1]{fontenc}    
\usepackage{hyperref}       
\usepackage{url}            
\usepackage{booktabs}       
\usepackage{amsfonts}       
\usepackage{nicefrac}       
\usepackage{microtype}      
\usepackage{graphicx}
\usepackage{subfig}
\usepackage{caption}

\title{Distributional Advantage Actor-Critic}

\author{
  Shangda Li \\
  Computer Science Department \\
  Carnegie Mellon University\\
  Pittsburgh, PA 15213 \\
  \texttt{shangdal@andrew.cmu.edu} \\
  \And
  Selina Bing \\
  Computer Science Department \\
  Carnegie Mellon University \\
  Pittsburgh, PA 15213 \\
  \texttt{zbing@andrew.cmu.edu} \\ 
  \And
  Steven Yang \\
  Computer Science Department \\
  Carnegie Mellon University \\
  Pittsburgh, PA 15213 \\
  \texttt{rujiay@andrew.cmu.edu}
 }

\begin{document}

\maketitle

\section*{Abstract}
In traditional reinforcement learning, an agent maximizes the reward collected during its interaction with the environment by approximating the optimal policy through the estimation of value functions. Typically, given a state $s$ and action $a$, the corresponding \textit{value} is the expected discounted sum of rewards. The optimal action is then chosen to be the action $a$ with the largest value estimated by value function. However, recent developments have shown both theoretical and experimental evidence of superior performance when value function is replaced with value distribution in context of deep Q learning [1]. In this paper, we develop a new algorithm that combines advantage actor-critic with value distribution estimated by quantile regression. We evaluated this new algorithm, termed Distributional Advantage Actor-Critic (DA2C or QR-A2C) on a variety of tasks, and observed it to achieve at least as good as baseline algorithms, and outperforming baseline in some tasks with smaller variance and increased stability.

\section*{Introduction}
In reinforcement learning, the \textit{value function} denotes the expected discounted reward when the agent is at state $s$. Typically, classic RL algorithms approximate action-value function $Q:\mathbb{S}\times\mathbb{A}\to\mathbb{R}$, the estimated discounted reward if action $a$ is taken at state $s$ by using the Bellman operator. Bellemare et al. have shown there exists a distributional equivalent of the Bellman operator [1], which has a theoretical guarantee of contraction under Wasserstein metric, an integral probability metric estimating distance between distributions [2]. Denote $Z^\pi$ as the random variable denoting the sum of discounted rewards under policy $\pi$, the equations below show the difference between Bellman operator and distributional Bellman operator:
\[Q^\pi(x,a) = \mathbb{E}[R(x,a)] + \gamma \mathbb{E}[Q^\pi(x',a')]\ , \quad Z^\pi(x,a) = R(x,a) + \gamma Z^\pi(x',a')\] 
A practical algorithm with theoretical equivalence to approximate distributional Bellman operator has been found recently by Dabney et al. using quantile regressions. The value distribution is approximated by using \textit{quantiles}: a fixed number of "atoms", or support points, with stochastically-adjustable locations are used to approximate the target distribution by minimizing Wasserstein distance. For a quantile distribution $Z_\theta\in Z_Q$,

\centerline{$Z_\theta(x,a) = \frac{1}{N}\sum_{i=1}^N\delta_{\theta_i(x,a)}\quad \delta_z $ denotes a Dirac at $z \in\mathbb{R}$}

The approximation of value distribution using quantile regressions is combined with Q-learning and evaluated on Atari 2600 environment, yielding significant improvements over baseline algorithms. [2] The advantage of using a value distribution is threefold: (1) the true value distribution contains more information than expected value, (2) increased robustness and stability during training due to learning of multimodality in value distribution, (3) effect of learning from nonstationary policy is mitigated. 

There are two major algorithms in the field of RL: Q-learning and policy gradient. Both approaches have their own drawbacks: Q-learning tends to have slower and poorer convergence, while policy gradient has the tendency to converge to local maxima and prone to have high variance. Advantage actor-critic (A2C) combines the benefits of both approaches by having an actor to approximate policy, and a critic to estimate value. [3] The algorithm described by Dabney et al. applied value distribution on Q-learning. In this paper, we propose to take a distributional approach on A2C algorithm: the underlying variance in values could be captured by the estimation of value distribution, thereby leading to more accurate and stable models. The actor, which approximates optimal policy, would remain unchanged. The critic would estimate the distribution of values directly using quantile regressions instead of simply estimating the expected sum of rewards.

\section*{Related Works}
\subsection*{Distributional Approach}
\centerline{Maximal form of Wasserstein metric: $\overline{d_p}(Z_1, Z_2) = \sup_{x,a}d_p(Z_1(x,a),Z_2(x,a))$}
Bellemare et al. has proved the distributional Bellman operator is a $\gamma$-contraction in a maximal form of the Wasserstein metric. The paper also devised the c51 algorithm that uses the idea of value distribution to solve Atari 2600 environments, yielding state-of-the-art solutions for many games. [1] However, the c51 algorithm leaves a theory-practice gap: instead of directly minimizing the Wasserstein metric as a loss, c51 first performs a heuristic projection step, then minimizes the Kullback-Leibler divergence between the projected update and the prediction. This is because when the loss is evaluated under Wasserstein metric, it could not be directly minimized using stochastic gradient methods. The estimation of value distribution is done by approximating variable probabilities $q_1, \ldots, q_N$ to fixed locations $z_1\leq \ldots \leq z_N$. In other words, $z_i$ is fixed to approximate $\Pr[X\leq z_i] = q_i$. [2] Since the release of c51 algorithm, there have been many evaluations of distributional q-learning, such as applying it on DQN, yielding state-of-the-art Rainbow optimizations [4], or applying it on policy gradient, yielding better performance than many baseline algorithms [5].

The recent quantile regression method devised by Dabney et al. closed the gap between theoretical and practical distributional approaches: the estimation of value distribution using quantile regressions (QR-DQN) is proven to be theoretically equivalent to distributional Bellman operator under Wasserstein metric. The use of quantile approximation has also been proven to have a theoretical guarantee of contraction [2]. Instead of fixing locations to parameterize adjustable probabilities, QR-DQN fixes probability quantiles $q_1, \ldots, q_N$ with adjustable locations $z_1,\ldots, z_N$, thereby providing more accurate prediction of low probability events at both ends of the distribution over returns. In addition, QR-DQN make use of the Wasserstein metric directly, instead of using Kullback-Leibler, which is a divergence and not a probability metric. As expected, QR-DQN has shown superior performance compared to its predecessor, the c51 algorithm, in a variety of Atari games. [2] It remains an open question as to whether quantile regression approximation methods could be used in other RL algorithms such as A2C, which is the topic of interest presented in this paper.

\subsection*{Advantage Actor-Critic}
One of the main drawbacks of on-policy learning algorithms is fundamentally unstable due to the strongly correlated data observed across timesteps. In 2016, Google DeepMind published a revolutionary paper on Asynchronous Advantage Actor-Critic, or A3C, which combines the benefits of both on- and off-policy algorithms by asynchronously executing multiple agents in parallel on different instances of environment, thereby decorrelating the agent's data and stabilizing the update process. In the algorithm, an actor learns the policy $\pi$ and the critic provides the baseline value $V(S_t)\approx b_t$, which is then used to scale the policy gradient by $A(a_t,s_t)=Q(a_t,s_t)-V(S_t)$. The use of $A(a_t,s_t)$, the \textit{advantage}, reduces variance by informing the agent of how much its prediction deviates from observed data. At each episode, $N$ actors (each existing in a separate worker thread) acts on a unique instance of the environment, and calculates the gradient update. The gradient updates then happen asynchronously at a central parameter server. Several implementation details were mentioned and were used in our paper: (1) \textit{Shared layers}: the performance were found to be the best when actor and critic share all their non-output layers; (2) \textit{Entropy on policy}: adding entropy $H$ on policy $\pi$ improves exploration and discourages premature convergence on local optima. Therefore, the objective function takes the form $\nabla_{\theta'}\log \pi(a_t|s_t;\theta')(R_t-V(s_t;\theta_v))+\beta \nabla_{\theta'}H(\pi(s_t;\theta'))$, with $\beta$ as the hyperparameter controlling the degree of entropy added to policy. [3]

Although A3C algorithm is a breakthrough that exceeded the performance of traditional RL algorithms, researchers wondered if the use of asynchrony is a necessity. A synchronous approach is then tested by OpenAI: the parameter server waits for all actors to finish before making a gradient update, which are averaged across all $N$ actors. The synchronous alternative, termed as A2C, in fact outperformed the original A3C algorithm. [6]

\section*{Methodology}
We evaluate DA2C's performance using a variety of OpenAI learning environments with different levels of difficulty. This is done to accurately learn the range of tasks DA2C is capable of performing. The performance is then compared with QR-DQN and regular A2C as baselines. We expect DA2C to outperform both baselines, as it is essentially a coalition of benefits of both.
\subsection*{Algorithms}
We would make use of three algorithms in this paper: 
\renewcommand{\labelitemi}{$\square$}
\begin{itemize}
\setlength\itemsep{0em}
\item \underline{Advantage Actor-Critic}: This algorithm is used as a baseline to evaluate whether there would be any improvements by modifying the value function to a value distribution estimated under quantile regression.
\item \underline{Distributional Q-Learning under Quantile Regression}: This algorithm is used as a baseline to evaluate whether substituting DQN with A2C would enhance performance. 
\item \underline{Distributional A2C}: This algorithm replaces critic's estimation of value function with an estimation of value distribution under quantile regression. Several modifications are tested against each other for optimal performance: (1) \textit{Shared Layers}: the original paper on A3C concluded that sharing of non-output layers between actor- and critic-model is superior to having separate models. We are curious as to whether this holds true under the distributional version of A2C. (2) \textit{Number of Atoms}: the original paper on c51 concluded that larger number of atoms always outperforms smaller number. We will evaluate whether this remains true when combined with A2C, although we hypothesize this is true.
\end{itemize}
\subsection*{Environments}
\renewcommand{\labelitemi}{$\square$}
\begin{itemize}
\setlength\itemsep{0em}
\item \underline{CartPole}: Classic control task with non-sparse reward, small state-space, and finite action-space. The goal for CartPole is to move the car left/right to ensure the pole stays upright.
\item \underline{MountainCar}: Classic control task with sparse reward. The task is difficult to solve due to the lack of reward feedback, and typically requires tweaks on entropy parameter and etc. 
\item \underline{LunarLander}: A moderate difficulty RL task with finite state- and action-space, in which a spaceship is expected to operate engine to move up, left, down, right to land in a designated position. The environment is different for each episode.
\item \underline{Atari}: Difficult RL task where the agent is expected to learn the environment from raw pixels. It has finite action- and state-space. The goal of Atari Assault is to maximize the player's score.
\end{itemize}
\subsection*{Atari Environment}
The default OpenAI Gym wrapper for the Atari environment has aspects of nondeterminism and requires some engineering to reach optimal performance. Therefore, many papers that reach the state-of-the-art performance ([7: nature and unreal/rainbow]) create wrappers directly using Arcade Learning Environment to set customized parameters for Atari environments. Instead of treating each frame as a separate state, we use frame-skipping and stack history frames to improve performance. Each state in our environment consists of 4 stacked frames (where the number of stacked frames is determined by hyperparameter \texttt{phi-length}). Each frame $f_t$ is element-wise max-pooled between $t$ and $t-1$ timestep. We perform maxpooling because Atari environments may sometimes only render objects once every two consecutive frames. Each frame in the state is 4 timesteps apart to ensure that each frame provides a distinct motion in the game. The stack of 4 frames provides the agent with a short history of motion, because the action chosen by agent depends on the prior sequence of frames (for instance, whether an object is going left or right cannot be determined by a single frame).
\begin{center}
Table 1: Atari Environment Setup \\
\begin{tabular}{||c|c|c||} 
\hline
Hyper-parameter & Value & Remarks \\
\hline
Gray-scale & True & \\
Downsampling & (84, 84) & With linear interpolation \\
Frame-skipping & 4 & \\
Action repetition & 4 & Disabled default stochastic action repetition \\
Color averaging & False & \\
Phi-length & 4 & Stacked frames per state \\ 
Reward clipping & [-1, 1] & \\
Terminal on loss of life & True & \\
Max frames per episode & 108K & \\
Minimal action set & True & \\
\hline
\end{tabular}
\end{center}

\section*{Results}
\subsection*{Classic Control Tasks}
The hyperparameters and network architectures are summarized in the table below. The model architecture is shared across QR-DQN, A2C, and QR-A2C. The same set of hyperparameters are applied uniformly to all three algorithms (if applicable) to ensure consistency and provide an accurate baseline for comparison.

\begin{center}
Table 2: Classic Control Task Setup \\
\begin{tabular}{||c|c||}
\hline
\hline
Model Layer & Activation \\
\hline 
Dense 64 & relu \\ 
Dense 64 & relu \\ 
Dense 64 & relu \\ 
\hline
\hline
Hyper-parameter & Value \\
\hline
Optimizer & Adam \\
Learning Rate & 7e-4 \\
N-Steps & 100 \\
Gamma & 0.99 \\
Num. Atoms & 16, 32, 64, 128 \\
Gradient Clip & 0.5 \\ 
Entropy Coef. $\beta$ & 0.01 \\
Value-Loss Coef. & 0.5 \\
Num. Workers & 1 \\
\hline
\end{tabular}
\end{center}
\underline{Shared and non-shared layers}: 
We compare the performance of shared non-output layers between actor and critic model and its non-shared equivalent under DA2C (referenced interchangeably in this text as QR-A2C). Other than the sharing of actor-critic model, all other hyperparameters are held constant, with 128 atoms being used to estimate the quantiles. Figure 1 shows the performance tested on CartPole environment. Although both seem to converge at a similar rate, we observe that shared model shows a much larger variance while converging, as evidenced in Figure 1b, and continues to show deviation from optimal policy after convergence (Figure 1a). 
\begin{figure}%
    \centering
    \subfloat[Shared/non-shared model comparison]{{\includegraphics[width=5cm]{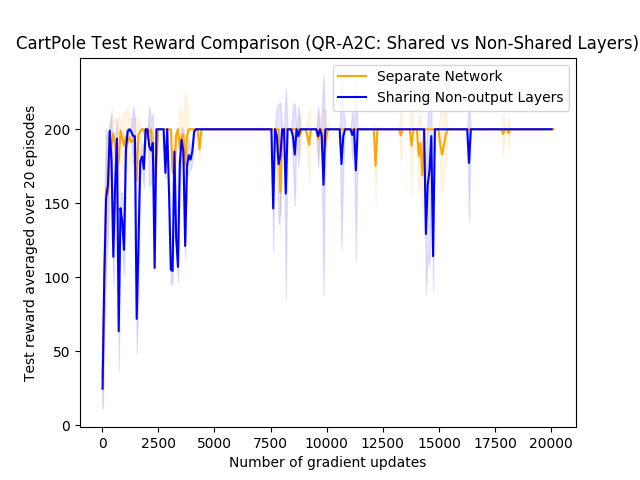} }}%
    \qquad
    \subfloat[Zoom in of comparison]{{\includegraphics[width=5cm]{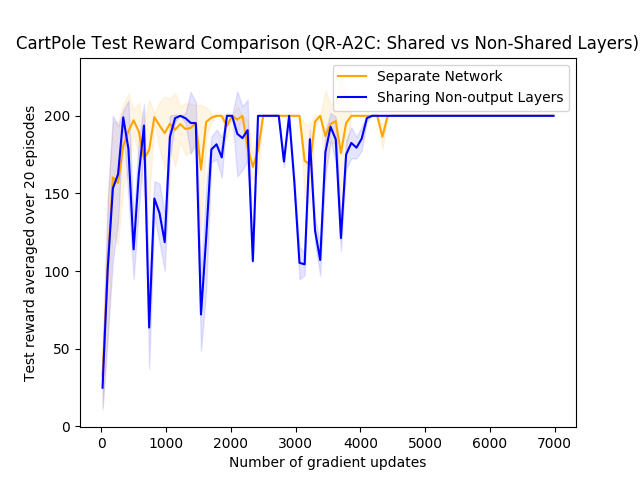} }}%
    \caption{Shared/Non-Shared Actor-Critic Model Comparison Tested on CartPole}%
    \label{fig:example}%
\end{figure}
\\ \\ 
\underline{Varying Atoms}: 
We compare the performance of varying atoms in QR-A2C algorithm on CartPole environment. 16, 32, 64, and 128 atoms were tested. The number of atoms is used to estimate quantiles for value distribution. We observe in Figure 2a that there is no significant difference between various atoms in CartPole environment. In Figure 2b, we observe that 64 atoms slightly outperformed 16 atoms in rate of convergence and variance. Adding number of atoms solves the environment faster, with higher test reward, and lower variance, as summarized in Table 3. We hypothesize this may be due to the simple nature of CartPole environment, therefore estimating smaller quantiles does not have large discernible impacts.
\begin{figure}%
    \centering
    \subfloat[16, 32, 64, 128 atoms]{{\includegraphics[width=5cm]{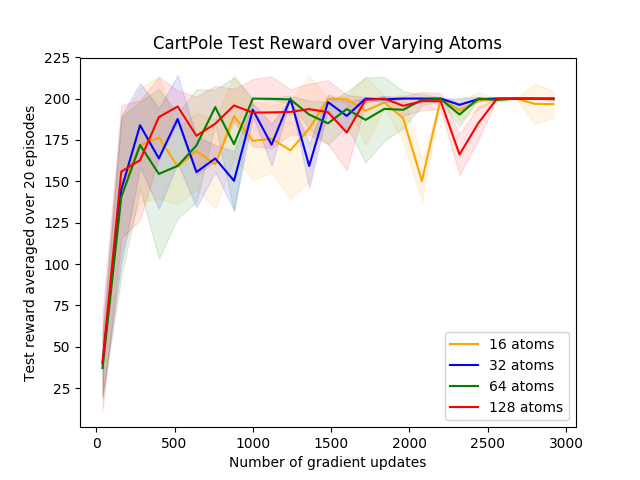} }}%
    \qquad
    \subfloat[16 and 64 atoms comparison]{{\includegraphics[width=5cm]{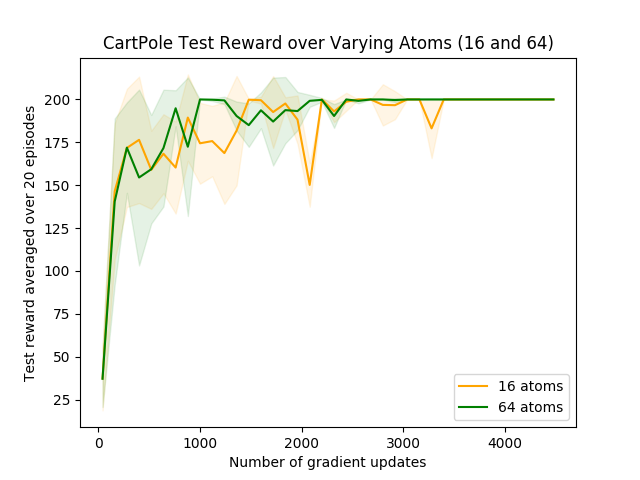} }}%
    \caption{QR-A2C over varying number of atoms tested on CartPole}%
    \label{fig:example}%
\end{figure}

\begin{center}
Table 3: Performance Comparison on Various Atoms \\
\begin{tabular}{||c|c|c|c||}
\hline
Num. Atoms & Num. Update to First Solve & Avg. Test Reward & Stddev. Test Reward \\
\hline 
16 & 960 & 195.55 & 13.76 \\
32 & 480 & 197.15 & 12.42 \\
64 & 720 & 200.00 & 0.00 \\
128 & 480 & 199.4 & 2.61 \\
\hline
\end{tabular}
\end{center}

\underline{CartPole Performance}: To evaluate the efficacy of our algorithm, we compare it with two baseline algorithms, A2C and QR-DQN. In Figure 3a, we can see that both A2C and QR-A2C significantly outperformed QR-DQN in terms of rate of convergence and variance. This could be attributed to the advantage of actor-critic-based algorithm over Q-learning. There does not seem to be any discernible difference between A2C and QR-A2C (Figure 3b) in terms of rate of convergence, although QR-A2C seem to show slightly better stability than A2C. We hypothesize that this could be attributed to the fact that CartPole environment's true value distribution is not complicated enough to observe difference in rate of convergence, but it does help to increase stability in training (as shown toward the end of Figure 1a: A2C diverged but QR-A2C, despite briefly diverging, quickly recovered to optimal policy.
\begin{figure}%
    \centering
    \subfloat[QR-A2C and baseline comparison]{{\includegraphics[width=5cm]{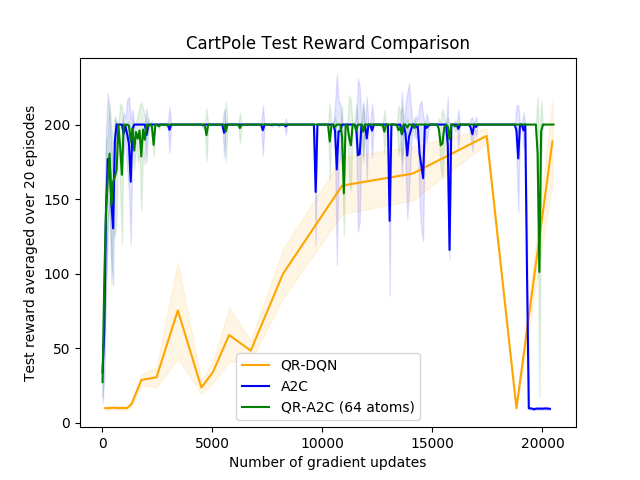} }}%
    \qquad
    \subfloat[A2C and QR-A2C comparison]{{\includegraphics[width=5cm]{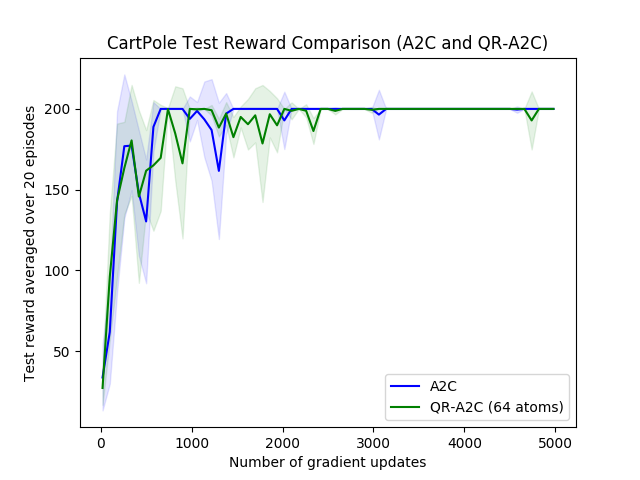} }}%
    \caption{CartPole Test Reward Comparison with Baseline Algorithms}%
    \label{fig:example}%
\end{figure}
\\ 
\underline{MountainCar Performance}: We were unable to solve MountainCar with QR-A2C or A2C, despite trying various entropy values and models. We believe this is a common problem for on-policy actor-critic algorithms, rather than being a deficient in QR-A2C. The policy could not be updated without seeing a reward, and without enough exploration, it is difficult to see any reward.
\newpage
\subsection*{Lunar Lander}
\begin{center}
Table 4: Lunar Lander Task Setup \\
\begin{tabular}{||c|c||}
\hline
Hyper-parameter & Value \\
\hline 
Learning rate & 1e-3 \\
N-step & 50 \\
Num. workers &  8 \\
\hline
\end{tabular}
\end{center}
Previously, we observed that QR-A2C showed little improvement over regular A2C under CartPole environment, and suspected this may be due to the true value distribution of CartPole being too simplistic. We therefore evaluate the performance of A2C and QR-A2C with various atoms on a more complicated environment, LunarLander. The model used is the same as CartPole environment. Hyperparameters are specified in Table 4 (those not specified remain unchanged from Table 2's values). We observe in Figure 4 that all QR-A2C instances, regardless of atom number, outperforms vanilla A2C algorithm. One remarkable observation is the superior performance of 32-atom to 64-atom to 120-atom. We suspect that the optimal number of atoms may be influenced by several factors, including the complexity of environment, as well as the tuning of other hyperparameters, which we did not have time to explore fully due to computational constraints.
\begin{figure}%
    \centering
    \subfloat[A2C and QR-A2C with various atoms]{{\includegraphics[width=5cm]{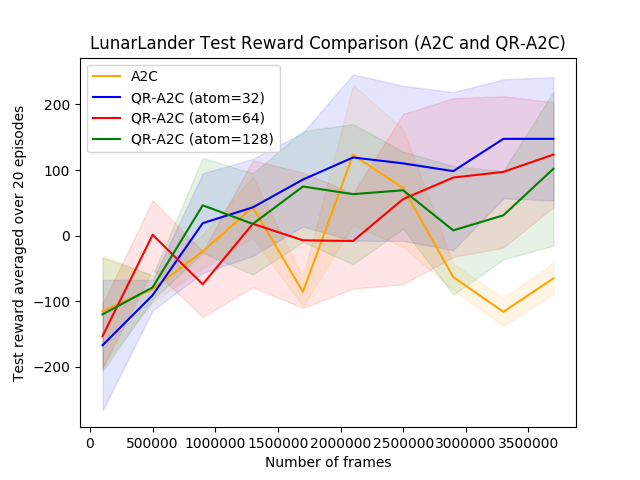} }}%
    \caption{LunarLander Test Reward Comparison with Baseline Algorithm}%
    \label{fig:example}%
\end{figure}
\newpage 
\subsection*{Atari}
\begin{center}
Table 5: Atari Task Setup \\
\begin{tabular}{||c|c|c||}
\hline
Model Layer & Activation & Others\\
\hline 
Conv2D & relu & stride=4, kernel=8, filter=32\\
Conv2D & relu & stride=2, kernel=4, filter=64 \\
Conv2D & relu & stride=2, kernel=4, filter=64 \\
Dense & relu & units=512 \\
\hline
\hline 
Hyper-parameter & Value & \\
\hline
Workers & 4 & \\
Num. Atoms & 64 & \\
N-step & 5 & \\
\hline
\end{tabular}
\end{center}
We evaluate on whether substituting value function with an estimation of value distribution using quantile regression would improve stability and rate of convergence on complicated environments such as Atari. The model and parameters used are specified in Table 5, in which the model is from DeepMind's 2015 paper on Atari. [7] We observe in Figure 5c that for A2C, shared model outperforms non-shared models; but for QR-A2C, non-shared model significantly outperforms shared models. We suspect this could be attributed to the complex nature of value distribution estimation, which renders it non-optimal to share layers. In shared models, A2C slightly outperforms QR-A2C, although A2C algorithm has a very high variance compared to its distributional counterpart (Figure 5a). In non-shared models, QR-A2C shows remarkable performance by outperforming both shared and non-shared A2C with a significant margin (Figure 5b, 5c), with similar variance to non-shared A2C, and achieves smaller variance than shared-A2C.
\begin{figure}%
    \centering
    \subfloat[Shared-Model Comparison]{{\includegraphics[width=5cm]{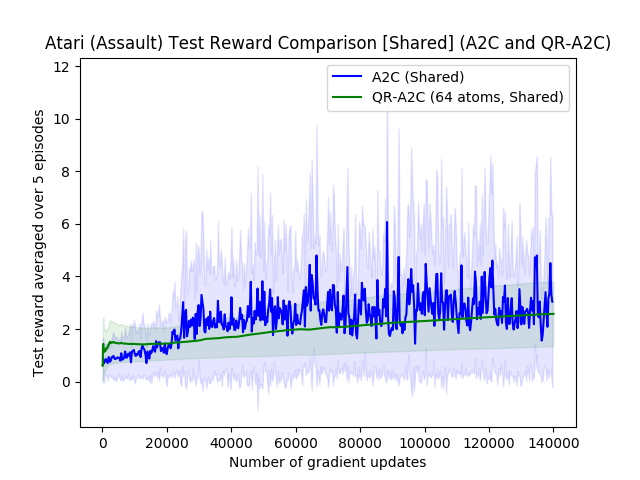} }}%
    \qquad
    \subfloat[Non-Shared Model Comparison]{{\includegraphics[width=5cm]{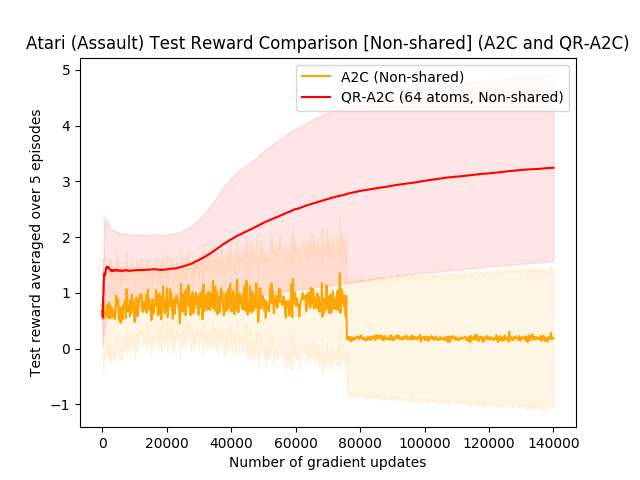} }}%
    \qquad
    \subfloat[A2C and QR-A2C comparison]{{\includegraphics[width=5cm]{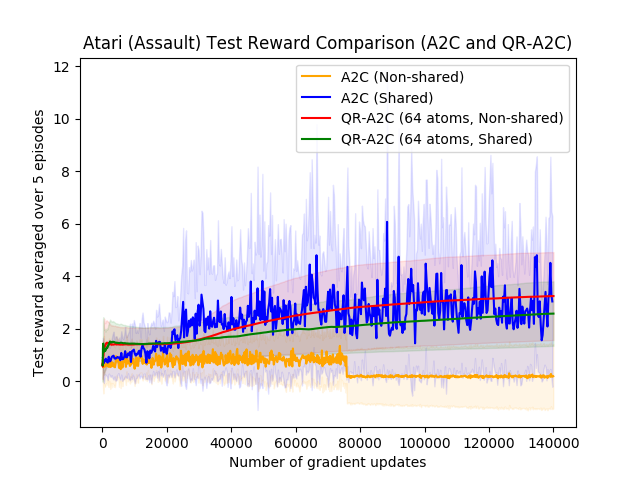} }}%
    \caption{Atari Assault Test Reward Comparison (QR-A2C and A2C)}%
    \label{fig:example}%
\end{figure}

\section*{Discussion}
\subsection*{Complexity of Environment}
One of the main observations we made during experimentation is that value distribution is more useful in more complex environments. We observed that QR-A2C showed little performance improvement in simple CartPole environment, but yields better result in Atari environment, where the underlying value distribution is more likely to be complicated and therefore benefits from quantile estimations of the distribution.
\subsection*{Number of Atoms}
Each atom in quantile distribution symbolizes a "support" point for the estimation of distribution. We hypothesized that the larger number of atoms would yield better performance due to a more accurate modeling of true distribution. Our experimental findings have mixed support for this viewpoint. Although in simple environments, there is only a marginal improvement when the number of atoms increased, in more complicated environments, we could observe a more discernible difference between different number of atoms. In LunarLander environment, increased number of atoms even seem to decrease the performance, which lead us to speculate the optimal number of atoms depend on the nature of the environment as well as other hyperparameters.
\subsection*{Reduce in Variance}
In two out of the three environments, we observed that compared to the use of value function, the value distribution reduces variance during training and also stabilizes the model once it converges. LunarLander environment is an outlier, in which the A2C algorithm yields lower variance than all of the QR-A2C variants. However, it is worth nothing that the A2C algorithm achieved a significantly lower score than the QR-A2C counterparts. We also observe that, generally speaking, increasing atoms would increase stability of model. The result supports theoretical reasoning that quantile approximation captures underlying value distribution.

\section*{References}
[1] Bellemare, Marc G., Will Dabney, and Rémi Munos. "A distributional perspective on reinforcement learning." \textit{arXiv preprint} arXiv:1707.06887 (2017). 

[2] Dabney, Will, et al. "Distributional Reinforcement Learning with Quantile Regression." \textit{arXiv preprint} arXiv:1710.10044 (2017). 

[3] Mnih, Volodymyr, et al. "Asynchronous methods for deep reinforcement learning." \textit{International Conference on Machine Learning}. 2016. 

[4] Hessel, Matteo, et al. "Rainbow: Combining Improvements in Deep Reinforcement Learning." \textit{arXiv preprint} arXiv:1710.02298 (2017). 

[5] Kumar, Aviral, and Govind Lahoti. “Distributional Stochastic Policy Gradients.” \textit{Indian Institute of Technology Bombay}, www.cse.iitb.ac.in/~aviral/DSPG.pdf. 

[6] OpenAI. “OpenAI Baselines: ACKTR \& A2C.” \textit{OpenAI Blog}, OpenAI Blog, 18 Aug. 2017, blog.openai.com/baselines-acktr-a2c/. 

[7] Dhariwal, Prafulla and Hesse, et al. "OpenAI Baselines" \textit{GitHub}, GitHub repository, 2017, https://github.com/openai/baselines/.

\end{document}